\documentclass{article}

\PassOptionsToPackage{numbers, compress}{natbib}
\usepackage[preprint]{neurips_2026}

\usepackage[utf8]{inputenc} % allow utf-8 input
\usepackage[T1]{fontenc}    % use 8-bit T1 fonts
\usepackage{url}            % simple URL typesetting
\usepackage{booktabs}       % professional-quality tables
\usepackage{amsfonts}       % blackboard math symbols
\usepackage{nicefrac}       % compact symbols for 1/2, etc.
\usepackage{microtype}      % microtypography
\usepackage{xcolor}         % colors

\usepackage{amsmath}
\usepackage{amssymb}
\usepackage{graphicx}
\usepackage{subcaption}
\usepackage{parskip}
\usepackage{tcolorbox}
\usepackage{hyperref}       % hyperlinks

\title{Meow-Omni 1: A Multimodal Large Language Model for Feline Ethology}

\author{%
  Jucheng Hu$^*$ \\
  University College London\\
  \texttt{jucheng.hu.20@ucl.ac.uk} 
  \And
  Zhangquan Chen$^*$ \\
  Tsinghua University \\
  \texttt{czq23@mails.tsinghua.edu.cn}
    \And
  Yulin Chen \\
  University College London \\
  \texttt{stephen.chen.22@ucl.ac.uk}
    \And
  Chengjie Hong \\
  University College London \\
  \texttt{zcabong@ucl.ac.uk}
    \And
  Liang Zhou \\
  University College London \\
  \texttt{zcablz0@ucl.ac.uk}
    \And
  Tairan Wang \\
  University College Londony \\
  \texttt{tairan.wang.22@ucl.ac.uk}
    \And
  Sifei Li \\
  University College London \\
  \texttt{zcabsl0@ucl.ac.uk}
    \And
  Giulio Zhu \\
  University College London \\
  \texttt{zcabgzh@ucl.ac.uk}
    \And
  Feng Zhou \\
  University College London \\
  \texttt{zcabfzh@ucl.ac.uk}
    \And
  Yiheng Zeng \\
  University College London \\
  \texttt{leo.zeng.22@ucl.ac.uk}
    \And
  Suorong Yang \\
  Nanjing University \\
  \texttt{sryang@smail.nju.edu.cn}
    \And
  Dongzhan Zhou \\
  Shanghai Artificial Intelligence Laboratory \\
  \texttt{zhoudongzhan@pjlab.org.cn}
}

\begin{document}
\footnotetext{$^*$Equal contribution.}

\maketitle

\begin{abstract}
Deciphering animal intent is a fundamental challenge in computational ethology, largely because of \textit{semantic aliasing}, the phenomenon where identical external signals (e.g., a cat's purr) correspond to radically different internal states depending on physiological context. Existing Multimodal Large Language Models (MLLMs) are blind to high‑frequency biological time‑series data, restricting them to superficial behavioural pattern matching rather than genuine latent‑state reasoning. To bridge this gap, we introduce \textbf{Meow‑Omni~1}, the first open‑source, quad‑modal MLLM purpose‑built for computational ethology. It natively fuses video, audio, and physiological time‑series streams with textual reasoning. Through targeted architectural adaptation, we integrate specialized scientific encoders into a unified backbone and formalize intent inference via physiologically grounded cross‑modal alignment. Evaluated on \textbf{MeowBench}, a novel, expert‑verified quad‑modal benchmark, Meow‑Omni~1 achieves state‑of‑the‑art intent‑recognition accuracy (71.16\%), substantially outperforming leading vision‑language and omni‑modal baselines. We release the complete open‑source pipeline including model weights, training framework, and the \textbf{Meow‑10K} dataset, to establish a scalable paradigm for inter‑species intent understanding and to advance foundation models toward real‑world veterinary diagnostics and wildlife conservation.
\end{abstract}

\section{Introduction}
The interpretation of animal behaviour has long been a cornerstone of veterinary 
medicine, wildlife conservation, and ethological research~\cite{jzbg2030031}. 
Yet deciphering the precise intentions of non‑verbal species remains extremely 
difficult because of the inherent ambiguity of their signals~\cite{Rendall_Owren_2013}. 
A feline purr, for instance, is frequently linked to contentment~\cite{article123132}, 
but it is equally documented as a self‑soothing mechanism during intense pain or 
respiratory distress~\cite{Merola2016}. We refer to this phenomenon as 
\textit{semantic aliasing}~\cite{Green_Angelaki_2010}: identical external signals 
can map to fundamentally different internal states depending on the physiological 
context. Relying on a single modality, such as visual cues or vocalisations alone, 
inevitably fails to resolve these critical ambiguities.

To bridge this gap, we introduce \textbf{Meow‑Omni~1}, a novel Multimodal Large 
Language Model (MLLM) specifically engineered to decode feline intent by natively 
fusing four data streams: text, video, audio, and biological signals (time‑series). 
Our design is driven by three foundational shifts in computational ethology:

\paragraph{\textbf{1) From Forecasting to Interpretation.}} 
Many existing AI approaches to animal behaviour focus on predictive sequence modelling, i.e., estimating the next video frame or the next pitch contour~\cite{Chen_2025,ntalampiras2019automatic,Wang_2022}. 
While statistically rigorous, such forecasts offer little insight into the 
animal's underlying state. Meow‑Omni~1 acts as a \textit{reasoning engine}, 
prioritising the decoding of latent intent (e.g., distinguishing occult pain 
from hunger) over simple pattern matching.

\paragraph{\textbf{2) The Necessity of LLM Reasoning.}} 
Decoding intention is not a straightforward classification task; it demands 
context‑aware synthesis. Large Language Models (LLMs) provide a framework for 
logical consistency and can draw on extensive knowledge about feline behaviour. 
By employing an LLM backbone, we obtain \textit{interpretable} inferences, 
enabling the model to articulate in natural language the association between a 
physiological spike and a behavioural display.

\paragraph{\textbf{3) Native Quad‑modal Grounding.}} 
To resolve the semantic aliasing defined above, the model must ground visual 
and auditory signals in physiological reality. Current MLLMs are typically 
blind to high‑frequency biological time‑series (TS) data. Meow‑Omni~1 is, to 
our knowledge, the first architecture in the ethological domain to unify 
internal biometrics directly within the linguistic embedding space. This 
enables the model to natively resolve ambiguities, for example, distinguishing 
playful aggression from territorial defence by correlating micro‑expressions 
with heart‑rate variability.

Technically, Meow‑Omni~1 is realised through careful architectural integration. 
We build on the multi‑modal backbone of MiniCPM‑o~4.5~\cite{openbmb2025minicpmo45} 
and augment it with specialised time‑series encoders from 
Intern‑S1~Pro~\cite{zou2026interns1proscientificmultimodalfoundation,bai2025interns1scientificmultimodalfoundation}. 
A custom Projection Layer maps the extracted biological features into the LLM's 
joint embedding space, giving the model a unified ``sense'' of the subject's 
physiological state.

To make this inference precise, we formalize animal intention via Pearl's 
structural causal models (Section~\ref{appendix:pf}), treating intention as the 
latent drive that would determine the animal's next action in a free‑choice 
environment. Meow‑Omni~1 is then trained to recover this latent intention from 
multimodal observations: the native fusion of video, audio, and physiological 
signals provides the necessary grounding, while expert annotations made under 
near‑free‑choice conditions supply the supervised target.

Although we concentrate initially on the domestic cat (\textit{Felis catus}), which is a 
species with a rich behavioural repertoire and accessible physiological 
monitoring, this work is designed as a scalable template for broader 
inter‑species communication. Our ultimate goal is to provide a practical tool 
for veterinary diagnosis and for the conservation of endangered wildlife.

We summarise our core contributions as follows:
\begin{itemize}
    \item \textbf{Meow‑Omni~1 Architecture:} The first native MLLM to unify 
    visual, auditory, and biological time‑series modalities for joint 
    behavioural reasoning.
    \item \textbf{The Meow‑10K Dataset:} A diverse, multi‑source dataset of 10 831 feline samples spanning varying modality combinations (video, audio, biometrics) with natural‑language descriptions.
    \item \textbf{MeowBench:} A novel, expert‑verified benchmark for evaluating 
    MLLMs on intention decoding and inter‑species reasoning.
    \item \textbf{Open Ethological AI Framework:} A complete, open‑source 
    pipeline including data, model weights, and training code that demonstrates how 
    multimodal foundation models can be adapted to non‑human species, paving 
    the way for applications in wildlife conservation.
\end{itemize}

\section{Related Work} \label{appendix:related_works}

\subsection{Traditional Methods in Animal Behaviour Interpretation}

The computational study of animal behaviour has evolved from manual ethogram coding to automated, single‑modality pipelines. These approaches fall broadly into three categories: acoustic analysis, postural recognition, and physiological monitoring.

\paragraph{Acoustic Analysis}
Significant effort has been devoted to the automatic classification of animal vocalizations. AVES~\cite{hagiwara2022avesanimalvocalizationencoder} leveraged massive unlabeled audio datasets to train a HuBERT‑based~\cite{hsu2021hubertselfsupervisedspeechrepresentation} self‑supervised model, which was subsequently fine‑tuned on downstream tasks from the BEANS benchmark~\cite{hagiwara2022beansbenchmarkanimalsounds}. This approach achieved performance comparable to fully supervised models, demonstrating strong transferability of learned audio representations. Building on this, Perch~2.0~\cite{burns2025perch20transferswhale} emerged as a bioacoustics foundation model, delivering state‑of‑the‑art results on marine mammal and underwater audio tasks.

Beyond universal models, species‑specific studies have targeted gibbons~\cite{inproceedingsgibbon}, birds~\cite{rauch2025maskedautoencoderslistenbirds}, and dolphins~\cite{semenzin2026dolphvec}. Specifically for felines, Ntalampiras~\cite{ntalampiras2019automatic} developed a framework that identifies contextual states such as waiting for food, isolation, or being brushed from cat meows, using advanced signal processing and pattern recognition techniques.

\paragraph{Visual Analysis}
Recent work has focused on developing universal visual representations that generalize across diverse animal taxa. Sun et al.~\cite{Sunbioarxivevf} first demonstrated that general‑purpose video foundation models can serve as effective feature extractors for animal behavior analysis, transferring well to unseen species. Extending this, the Universal Action Space (UAS)~\cite{chang2026universalactionspacegeneral} enables cross‑species behavior analysis without requiring backbone fine‑tuning. For few‑shot adaptation, UniAP~\cite{sun2023uniapuniversalanimalperception} employs prompt‑based learning to support cross‑species perception in tasks like pose estimation and segmentation. In postural analysis, SuperAnimal~\cite{Ye2024NC} provides pretrained foundation models, drastically reducing the need for labeled data.

Focusing on specific species, SIPEC~\cite{Marks_2022} offers an integrated pipeline for segmentation in socially interacting primates and mice. As for felines, vision‑based affect detection has centred on facial landmarks. The Feline Grimace Scale has been digitized using DeepLabCut and CNNs to detect pain indicators. For instance, Feighelstein et al.~\cite{Feighelstein2023} employed a multi‑view setup to classify feline pain with high sensitivity. However, these systems often degrade in low‑light or occluded conditions. In such scenarios, audio or biometric modalities can provide critical complementary signals.

\paragraph{Temporal Modeling}
Time‑series (TS) data in animal behavior research typically originate from raw wearable sensor streams (e.g., accelerometers) and kinematic trajectories derived from pose estimation. Across species, these sequential signals are commonly mapped to behavioral states using classical classifiers~\cite{FANG2021105863,10278344}. With the maturation of deep learning, One‑dimensional Convolutional Neural Networks (1D‑CNNs) and Long Short‑Term Memory (LSTM) networks are widely used to process both raw sensor windows and keypoint trajectories~\cite{Fazzari2024}. For multi‑sensor fusion, hybrid 1D‑CNN/LSTM pipelines remain dominant~\cite{ARABLOUEI2023107707,9967209,10028752}.

In the feline domain, TS modeling remains largely limited to discrete physical actions. Recent work by \cite{s24237436} applied 1D‑CNN and LSTM architectures to recognize simple behaviours like walking, grooming, and eating in cats using IMU data.

\subsection{The Rise of MLLMs}

While traditional deep learning excels at isolated pattern recognition, it fundamentally lacks the semantic reasoning required to infer complex intent. The artificial intelligence landscape has thus undergone a paradigm shift, transitioning from modular, pipeline‑based architectures (which rely on “late fusion” of predictions) to unified, end‑to‑end MLLMs capable of native cross‑modal integration.

\paragraph{Leading Foundation Models}
Recent advancements have yielded highly capable Vision‑Language Models (VLMs) and Omni‑modal models. Proprietary architectures, such as Claude Opus~4.7~\cite{anthropic2026claudeopus47} and Gemini~3.1~Pro~\cite{googledeepmind2026gemini31pro}, alongside open‑source leaders like Qwen3.5‑397B‑A17B~\cite{qwen3.5} and Qwen3‑Omni‑30B‑A3B~\cite{xu2025qwen3omnitechnicalreport}, demonstrate remarkable capabilities in integrating text, high‑resolution vision, and audio. These models achieve instant, human‑like sensory integration. However, their architectural focus remains predominantly constrained to human‑centric modalities. They excel in linguistic speech understanding while treating non‑linguistic signals as background noise.

\paragraph{Scientific and TS Multimodality}
Extending MLLMs into non‑linguistic and numerical domains represents a crucial new frontier. Notably, models such as Intern‑S1‑Pro~\cite{zou2026interns1proscientificmultimodalfoundation} have pioneered the integration of continuous TS modalities natively into the LLM framework. Unlike conventional models that treat numerical data as flat text, Intern‑S1‑Pro employs specialized temporal encoders to process raw, high‑frequency scientific sensor readings and physiological data, grounding continuous 1D signals within a semantic embedding space.

\paragraph{Existing Constraints in Computational Ethology}
Despite the immense power of these foundation models, they face three critical limitations when applied to animal behavior:
\begin{enumerate}
    \item \textbf{Modality Blindness and Integration:} While Intern-S1-Pro supports TS data, and models like Gemini or Qwen excel in audio-visual streams, no existing general-purpose MLLM natively co-embeds high-frequency biological TS data (e.g., IMU/ECG) \textit{alongside} both audio and video. Current architectures treat physiological readings as disparate streams.
    \item \textbf{The Symbol Grounding Gap:} Current models lack the sensory-motor grounding to correlate a real-world physiological spike (e.g., a 50Hz acceleration burst) with a specific behavioral micro-expression or non-linguistic feline vocalization.
    \item \textbf{Domain Bias:} Trained overwhelmingly on human data, these models frequently misinterpret feline communication such as subtle frequency shifts, pupil dilation, and nuanced tail positioning, as generic "animal noise."
\end{enumerate}

\subsection{Benchmarks and Evaluation Vacuum}
Evaluation frameworks for animal intelligence are remarkably sparse. While general benchmarks like Animal‑Bench~\cite{jing2024animalbench} and MammalNet~\cite{chen2023mammalnetlargescalevideobenchmark} exist, they focus predominantly on species recognition or basic action labeling (e.g., “running”). There is currently no standardized framework for \textit{intent inference} (distinguishing “defensive aggression” from a “playful swipe”). Meow‑Omni~1 addresses this by introducing MeowBench, the first benchmark specifically designed to test cross‑modal reasoning between biometrics, vision, and audio.

% \subsection{Meow-Omni 1: Solving the Limitation}
% Meow-Omni 1 performs "model surgery" to bridge the sensory divide. By integrating the specialized TS encoder of Intern-S1 Pro \cite{zou2026interns1proscientificmultimodalfoundation} into the omni-modal backbone of MiniCPM-o 4.5, we enable the model to process 50Hz biological signals natively alongside audio and video. This integration allows for unified reasoning, resolving the ambiguity of an isolated action by correlating it directly with corresponding heart rate variability and kinematic markers.

\section{Problem Formulation} \label{appendix:pf}

To provide a rigorous basis for the Meow-Omni 1 architecture, we move beyond heuristic labeling of behavior toward a formal computational framework. This section defines our theoretical stance on feline intent and formalizes the multimodal mapping task.

\subsection{Formal Definition of Animal Intention}
In classical ethology, intention is often treated as a subjective cognitive state, complicating quantitative analysis. Modern computational ethology, however, increasingly models behavior as a sequence of observable actions governed by unobservable (latent) internal states \cite{Pereira2020}. Concurrently, the Active Inference framework in neuroscience posits that biological systems act to minimize the divergence between their current state and a predicted, optimal future goal state \cite{Parr2022}.

We synthesize these perspectives using formal causal inference, specifically Pearl's structural causal models \cite{Pearl2019}. We formally define \textit{Animal Intention} ($\mathcal{I}$) as the \textbf{latent biological and cognitive drive that maximizes the probability of a specific future action, under the hypothetical intervention of placing the animal in an unconstrained environment}. 

Let an animal's observable history up to time $t$ be a multimodal sequence $\mathcal{H}_t = \{V_{1:t}, A_{1:t}, B_{1:t}\}$, representing Vision, Audio, and Biometric TS data, respectively. In a constrained environment $\mathcal{E}_{c}$ (e.g., a cage, a clinical setting, or the presence of a handler), the actual next action $a_{t+1}$ may be suppressed or altered by external confounders.

To decouple the animal's true intent from these environmental constraints, we model intention as the latent variable dictating the optimal next action $a^*_{t+1}$ under a forced intervention. Specifically, setting the environment to a free-choice state, denoted by Pearl's $do$-operator as $do(\mathcal{E}_{free})$. We formalize this as:
\begin{equation}
    \mathcal{I} = \arg\max_{i \in \mathbb{I}} P(a^*_{t+1} \mid \mathcal{H}_t, do(\mathcal{E}_{free}), i)
\end{equation}
where $\mathbb{I}$ represents a discrete set of semantic intent categories (e.g., \textit{Seeking Food, Defensive Evasion, Self-Soothing Pain Relief}). The introduction of the $do$-operator is critical: it ensures we are defining the causal effect of the internal state $i$ on the proposed action $a^*_{t+1}$, independent of the observational biases and physical limitations introduced by the current setting $\mathcal{E}_c$.

\subsection{The Multimodal Inference Task}
The objective of Meow-Omni 1 is to approximate the function $f_\theta: \mathcal{H}_t \to \mathbb{I}$. Unlike traditional action recognition models that map $\mathcal{H}_t$ directly to an observational label $y_t$ (e.g., "jumping"), our model must learn a joint distribution that accounts for the hidden physiological state driving the causal graph.

Given the multimodal input $\mathcal{H}_t$ recorded in a constrained context $\mathcal{E}_c$, the task is to maximize the log-likelihood of the correct intent $\mathcal{I}$ by aligning external modalities (video, audio) with the internal biometric state:
\begin{equation}
    \mathcal{L}(\theta) = \sum_{(\mathcal{H}_t, \mathcal{I}) \in \mathcal{D}} \log P( \mathcal{I} \mid \text{Encoder}(\mathcal{H}_t); \theta)
\end{equation}
By natively grounding the model in biometric data $B_{1:t}$, we enable the system to resolve the semantic aliasing, i.e., scenarios where the same visual or auditory signal (e.g., a purr) corresponds to vastly different values of $\mathcal{I}$ (e.g., contentment vs. pain) depending on the underlying physiological markers.

\begin{figure}[htbp]
    \centering
    \includegraphics[width=1\textwidth]{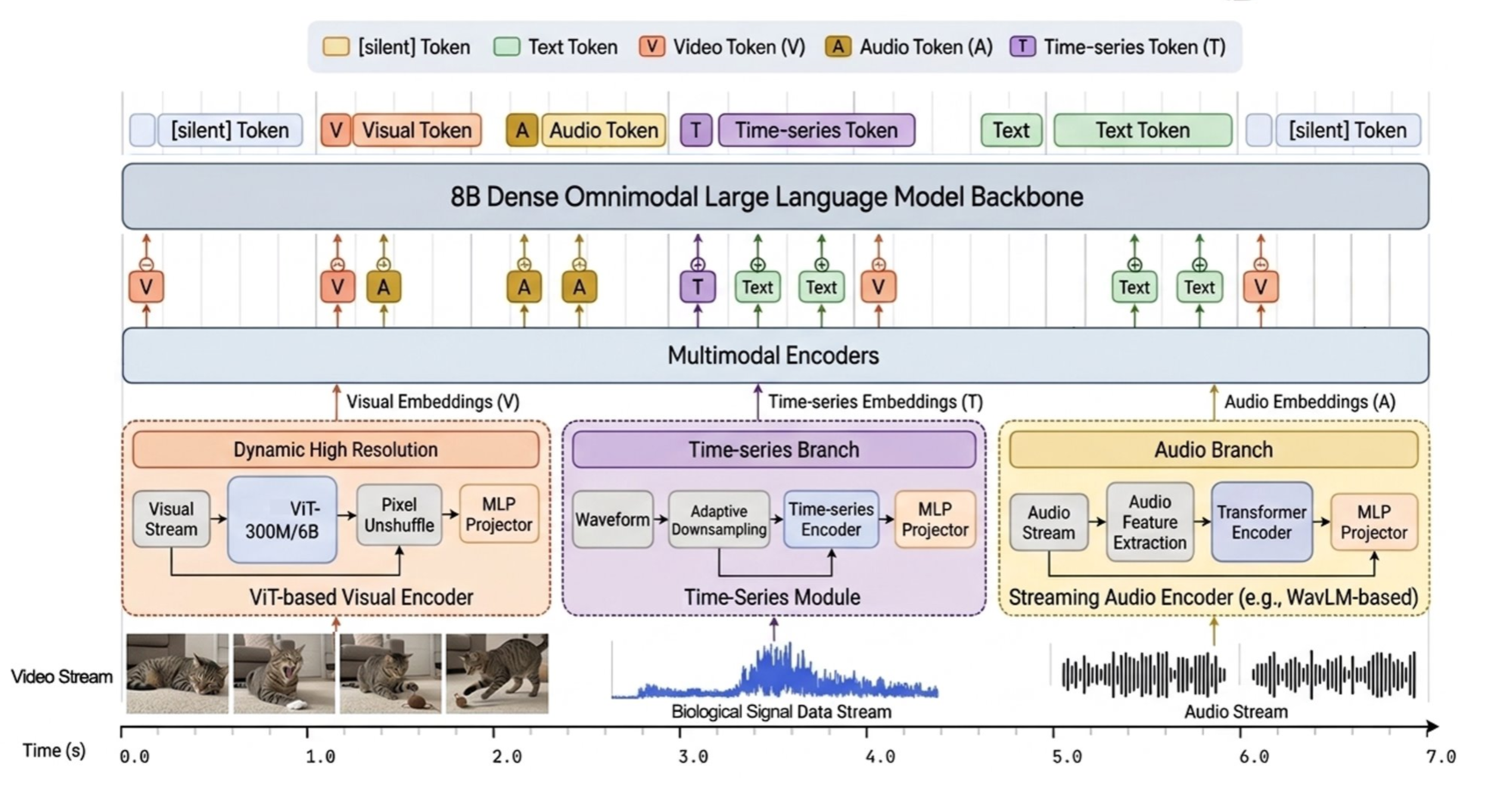}
    \caption{Meow-Omni 1 Model Architecture.}
    \label{fig:arch}
\end{figure}

\section{Methods}
\label{sec:methods}

The development of Meow-Omni~1 follows a multi-stage approach encompassing specialized model surgery, a novel temporal labeling strategy, and a rigorous alignment-specialization training pipeline.

\subsection{Architecture: Model Surgery and Tokenizer Expansion}
We build Meow-Omni~1 upon the MiniCPM-o backbone, performing a series of architectural ``transplantations'' to accommodate feline-specific biometrics. The overall architecture is illustrated in Figure~\ref{fig:arch}. 

\textbf{1) Vocabulary Expansion.} We extend the MiniCPM tokenizer with three unique control tokens: \texttt{<|ts\_start|>}, \texttt{<|ts\_unit|>}, and \texttt{<|ts\_end|>}. To handle varying lengths of time-series (TS) data, a \texttt{MeowOmniProcessor} dynamically expands the \texttt{<|ts\_unit|>} placeholder to match the number of hidden states produced by the encoder.

\textbf{2) Model Surgery.} We implement the \texttt{MeowOmni1PreTrainedModel} class, utilizing a customized \texttt{MeowOmni1Config}. This configuration integrates a dedicated TS encoder and a linear projector adapted from Intern‑S1‑Pro. We perform surgery on the base LLM's embedding layer, resizing it to accommodate the new TS tokens and ensuring seamless modal integration within the causal transformer block.

\textbf{3) Multimodal Forward Pass.} The primary class, \texttt{MeowOmni1ForCausalLM}, customizes the forward pass to accept \(N\)-dimensional TS tensors alongside standard vision and language inputs. TS embeddings are projected into the LLM's hidden dimension and interleaved with visual and linguistic embeddings, forming a unified multimodal context. The model is designed to process variable-length sequences that may contain any subset of the available modalities; absent modalities are simply omitted from the input sequence.

\subsection{Training Pipeline: Alignment and Specialization}
\label{sec:training}
The training process is divided into two distinct phases to ensure modal stability.

\textbf{Stage 1: Projector Alignment.} We pre-train the TS projector using 383,853 labeled TS samples (described in Appendix~\ref{appendix:DPP}). During this stage, the LLM backbone and the TS encoder are frozen; only the projector is updated to map biometric features into the linguistic latent space. An early-stopping strategy with 1-epoch patience is employed, yielding an optimal alignment checkpoint after 2 epochs, which we denote as \textit{Meow-Omni~1 Aligned}.

\textbf{Stage 2: Multimodal Specialization.} We fine-tune using the Meow‑10K dataset, which comprises 10,831 high-quality samples with varying modality combinations (A/V/TS, A/V, V/TS, A/TS, single-modality, etc.). All encoders and projectors are frozen, and only the LLM backbone is updated. Missing modalities are absent from the input sequence (no placeholder tokens are used), allowing the model to learn to reason from any subset of the available streams. Using the same 1-epoch patience, the final \textit{Meow-Omni~1} model is obtained after 2 epochs of fine-tuning.

\subsection{Dataset Generation}
\label{sec:dataset_generation}
The Meow‑10K training set is assembled from multiple pipelines; full details are provided in Appendix~\ref{appendix:DPP}. Each sample ultimately carries a natural-language query-response pair and, where applicable, a label from a unified 30-class intention taxonomy (listed in Appendix~\ref{appedix:label_list}). The main labeling strategies are summarized below:

\begin{itemize}
    \item \textbf{Time‑Series (TS) Data.} We employ a Next‑Behaviour Prediction (NBP) labeling strategy: for a fixed-length accelerometer window, the target is the future behaviour label taken from the original 30-class intentions provided by the source datasets. Intermediate transient movements are filtered out to retain only semantically stable intention states.
    \item \textbf{Video‑Only Clips.} A VLM-based pipeline detects action onsets and generates natural-language action captions. Each caption is subsequently mapped to one of the 30 intention classes by an LLM; the mapping is verified by the same expert group that curated MeowBench.
    \item \textbf{Audio‑Only Clips.} Standalone audio recordings are captioned with rich behavioural descriptions generated by a text-only LLM expansion of the original short captions. No pre-assigned intention label is given; the model is trained to produce a descriptive caption from the audio signal.
    \item \textbf{Synchronized Audio-Video Pairs.} Clips derived from AudioSet inherit their temporal alignment from the source videos. A VLM jointly analyzes audio and video to produce an audio-focused caption and a probability distribution over the 30 intention classes, which serves as the supervision target.
    \item \textbf{Synthetic Quad‑Modal Samples.} A small number of samples combine video-audio pairs with TS data that share the same intention label, following the same protocol as MeowBench (described next). These samples are verified by the expert reviewers and are included to teach the model joint tri‑modal reasoning.
\end{itemize}

After balancing modality representation (by subsampling the abundant TS‑only data to 2,000 samples), the final Meow‑10K dataset contains 10,831 training samples.

\subsection{MeowBench: Intent‑Matched Synthesis and Evaluation}
\label{sec:meowbench}
To evaluate the model's multi‑modal reasoning, we curated \textbf{MeowBench}, a held‑out evaluation suite. Since naturally synchronized quad‑modal datasets do not exist, we synthesized samples by matching unimodal data that share the same intention label. Specifically, for a given intention, we pair a video-audio clip (recorded from the same individual cat) with a TS sample sourced from a different session but annotated with the identical intention. To ensure physical plausibility, the synthesized combinations were reviewed by eight Professional Feline Ethologists. The experts worked in three groups (sizes 3, 3, and 2), each group jointly evaluating roughly one third of the 645 initial candidates and reaching an internal consensus. This verification process refined the set from 645 to 527 high-fidelity samples. Each sample is then converted into a Multiple Choice Question (MCQ): the correct intention serves as the answer, and three randomly sampled distractors from the broader 30-class label set are used as alternatives, testing the model's discriminative accuracy.
\section{Experiments}

In this section, we describe the experimental setup, baseline models, and ablation strategies used to validate Meow‑Omni~1’s capability to decode feline intention from quad‑modal inputs.

\subsection{Baselines for Comparison} \label{app:baselines_intro}
Given that Meow‑Omni~1 is the first MLLM to process quad‑modal feline data (Text, Video, Audio, and Biometrics), we compare our model against state‑of‑the‑art (SOTA) unimodal and bi‑modal systems to establish performance benchmarks.

\begin{enumerate}
    \item \textbf{Unimodal Baselines:} 
    \begin{itemize}
        \item \textit{Vision:} We utilize \textbf{Qwen3.5‑122B‑A10B}, the leading open‑source Vision‑Language Model at the time of this study, as a strong zero‑shot baseline for visual intent inference.
        \item \textit{Audio:} We compare against the specialized framework by \cite{ntalampiras2019automatic}, which employs signal processing and pattern recognition for context‑aware cat vocalization classification.
        \item \textit{TS:} We benchmark against the feline behavior recognition model proposed by \cite{Chen_2025}, which uses deep learning for IMU‑based behavior classification.
    \end{itemize}
    \item \textbf{Bi‑modal Baselines:} We evaluate \textbf{Qwen3.5‑Omni‑Plus}, a leading omni‑modal foundation model, using various two‑modal combinations (e.g., Video + Audio) from our MeowBench suite.
    \item \textbf{Quad‑modal Comparison:} As Meow‑Omni~1 is the first model to natively co‑embed high‑frequency biometrics with audio‑visual streams, no existing quad‑modal baseline is available for direct comparison.
\end{enumerate}

Detailed descriptions of model architectures, preprocessing pipelines, and training hyperparameters for each modality are provided in Appendix~\ref{appendix:baselines}.

\subsection{Evaluation Metrics}
All models are evaluated on the \textbf{MeowBench} MCQ suite (527 expert‑verified samples). We report \textbf{Top‑1 Accuracy} for intention matching. % (Macro‑F1 removed because it was never reported; if you wish to keep it, add the corresponding column to Tables 1 and 2.)

\subsection{Uncertainty Quantification via Temperature Sampling}
While standard evaluation metrics capture the model’s accuracy on congruent data, real‑world veterinary and ethological applications demand robust Uncertainty Quantification (UQ). Traditional LLM softmax outputs are frequently miscalibrated and entangle aleatoric (data) and epistemic (model) uncertainty. To evaluate Meow‑Omni~1’s ability to recognize inherently ambiguous or contradictory multimodal signals, we employ a sampling‑based predictive entropy approach.

\subsubsection{Conflict Dataset}
We design a controlled inference experiment to probe the model’s confidence under adversarial conditions. We evaluate the model on two distinct subsets. For the control group (congruent), 50 randomly sampled instances from the MeowBench suite where Video, Audio, and TS modalities align perfectly toward a single ground‑truth intent. For the test group (conflict), 50 synthesized adversarial instances. In this subset, the Video and Audio modalities are paired to indicate Intention~A (e.g., “Contentment”), while the injected TS biometrics are intentionally mismatched to indicate Intention~B (e.g., “Occult Pain”).

\subsubsection{Predictive Entropy Formulation}
Rather than relying on a single deterministic decoding pass, we force the model to reveal its internal uncertainty by sampling from its output distribution multiple times. For each instance in both the Control and Conflict groups, we execute $N = 10$ independent stochastic forward passes using a generation temperature of $T = 0.7$.

Let $C$ represent the set of all unique semantic intention categories generated across the $N$ samples for a given multimodal sequence $\mathcal{H}_t$. We calculate the empirical probability $\hat{p}(c)$ of the model predicting intention $c$. The uncertainty of the model’s inference is then quantified using Predictive Shannon Entropy:
\begin{equation}
    H(\mathcal{H}_t) = - \sum_{c \in C} \hat{p}(c) \log \hat{p}(c)
\end{equation}
A low entropy score ($H \approx 0$) indicates high confidence and modal agreement. Conversely, a high entropy score indicates that the model has detected the injected ambiguity, correctly distributing its probability mass across the competing modalities (aleatoric uncertainty) or yielding a uniform distribution due to out‑of‑distribution inputs (epistemic uncertainty).

\subsection{Ablation Study: Modality Masking}
\label{app:ablation_setup}
To quantify the synergistic effect of our quad‑modal architecture, we perform an extensive ablation study using a \textbf{modality masking} strategy. We evaluate the Meow‑Omni~1 final checkpoint on the MeowBench suite under 3 masking conditions, including \textbf{Unimodal Masking:} Masking two out of three sensory modalities to observe the model’s reliance on single streams (e.g., evaluating on $V$ only while masking $A$ and $TS$).
\textbf{Bi‑modal Masking:} Masking one modality (e.g., $V+A$ with $TS$ masked) to measure how the addition of biometrics resolves visual or auditory ambiguities.
\textbf{Full Integration:} Utilizing all three sensory modalities (Video, Audio, TS) together with the linguistic prompt to establish the performance ceiling.

\section{Results}

In this section, we present a quantitative evaluation of Meow‑Omni~1. We first compare our architecture against SOTA baselines, followed by a comprehensive ablation study and an analysis of the model’s uncertainty under signal conflict.

% Move Table 1 here so that it is discussed immediately after its reference.
\begin{table}[t]  
\centering
\caption{Comparison of Meow‑Omni~1 against SOTA baselines on MeowBench.}
\label{tab:baselines}
\begin{tabular}{lcccc}
\hline
\textbf{Model} & \textbf{Vision} & \textbf{Audio} & \textbf{TS} & \textbf{Accuracy (\%)} \\ \hline
Acoustic SOTA \cite{ntalampiras2019automatic} & & \checkmark & & 36.86\% \\
Video SOTA (Qwen3.5‑122B) & \checkmark & & & 61.95\% \\
TS SOTA \cite{Chen_2025} & & & \checkmark & 48.98\% \\ \hline
Qwen3.5‑Omni‑Plus (V+A) & \checkmark & \checkmark & & 65.36\% \\
Qwen3.5‑Omni‑Plus (V+TS) & \checkmark & & \checkmark & 66.21\% \\
Qwen3.5‑Omni‑Plus (TS+A) & & \checkmark & \checkmark & 42.15\% \\
Qwen3.5‑Omni‑Plus (V+A+TS) & \checkmark & \checkmark & \checkmark & 66.89\% \\ \hline
\textbf{Meow‑Omni~1 (Ours)} & \checkmark & \checkmark & \checkmark & \textbf{71.16\%} \\ \hline
\end{tabular} \vspace{-1em}
\end{table}

\subsection{Main Results and Benchmark Comparison}
As shown in Table~\ref{tab:baselines}, Meow‑Omni~1 achieves a Top‑1 accuracy of \textbf{71.16\%}, outperforming all unimodal and multimodal baselines. Notably, our model surpasses the leading general‑purpose omni‑modal baseline (Qwen3.5‑Omni‑Plus with all three modalities, 66.89\%), which treats non‑linguistic signals with generic encoders. This 4.3‑percentage‑point improvement validates our hypothesis that native, high‑frequency biological grounding is essential for resolving behavioural intent.

\begin{table}[h] 
\centering \vspace{-1em}
\caption{Ablation study of modality contributions within Meow‑Omni~1.}
\label{tab:ablation}
\begin{tabular}{lcccc}
\hline
\textbf{Modality Configuration} & \textbf{Vision} & \textbf{Audio} & \textbf{TS} & \textbf{Accuracy (\%)} \\ \hline
Unimodal (Audio) & & \checkmark & & 51.88\% \\
Unimodal (Vision) & \checkmark & & & 69.97\% \\
Unimodal (TS) & & & \checkmark & 55.63\% \\ \hline
Bi‑modal (V+A) & \checkmark & \checkmark & & 68.43\% \\
Bi‑modal (V+TS) & \checkmark & & \checkmark & 70.82\% \\
Bi‑modal (TS+A) & & \checkmark & \checkmark & 54.95\% \\ \hline
\textbf{Full Quad‑modal (V+A+TS)} & \checkmark & \checkmark & \checkmark & \textbf{71.16\%} \\ \hline
\end{tabular} \vspace{-0.5em}
\end{table}

\subsection{Ablation Study: Modal Synergy}
To quantify the contribution of each modality to the final reasoning process, we performed a modality‑masking ablation study on the final model checkpoint. The results are summarized in Table~\ref{tab:ablation}.

The ablation reveals that adding biological TS data yields the most notable gain among the modalities. In the baseline comparison (Table~\ref{tab:baselines}), the jump from the video‑only SOTA baseline (61.95\%) to the bi‑modal V+TS omni‑baseline (66.21\%) is substantial, and further addition of audio leads to the top performance of Meow‑Omni~1 (71.16\%). Within our own architecture, the unimodal Vision result is already strong (69.97\%), and the inclusion of TS (V+TS: 70.82\%) provides a moderate boost, confirming that biological markers help resolve the “semantic aliasing” often found in purely visual or auditory feline displays. The full quad‑modal integration consistently reaches the highest accuracy.

\begin{table}[t]
\centering
\caption{Predictive entropy ($H$) results for congruent vs.\ conflicting modalities (preliminary results).}
\label{tab:uncertainty}
\begin{tabular}{lcc}
\hline
\textbf{Dataset Group} & \textbf{Avg.\ Entropy (bits)} & \textbf{Primary Outcome} \\ \hline
Congruent (Matching) & 1.28 & High Confidence / Agreement \\
Conflict (Mismatched) & 3.15 & High Uncertainty / Divergence \\ \hline
\end{tabular} \vspace{-1em}
\end{table}

\subsection{Uncertainty Quantification Analysis}
Finally, we analysed the model’s reliability using the temperature‑sampling method ($N=10$, $T=0.7$) described above. We compared the predictive entropy $H$ between the congruent control group and the synthesized conflict dataset.

The results (Table~\ref{tab:uncertainty}) indicate a sharp divergence in entropy. While the model remains decisive on congruent data ($H = 1.28$~bits), it exhibits significantly higher entropy on conflicting samples ($H = 3.15$~bits). This confirms that Meow‑Omni~1 does not simply default to the strongest modality (Vision) but genuinely attempts to reconcile contradictory biological data, resulting in a quantifiable “hesitation” that can be used to alert human observers to ambiguous states.

\section{Discussion}

The empirical evaluation of Meow‑Omni~1 confirms our core hypothesis: native integration of high‑frequency biological time‑series data alongside audio‑visual streams significantly enhances the capacity of foundation models to decode non‑verbal intent.

\subsection{Resolving Semantic Aliasing}
The full quad‑modal architecture achieves 71.16\% accuracy, outperforming the best general‑purpose omni‑modal baseline (Qwen3.5‑Omni‑Plus with video, audio, and a textual summary of biosignals, 66.89\%). This 4.3‑point margin underscores the limitations of relying solely on external observational data, even when some physiological information is provided as text.

Our ablation study (Table~\ref{tab:ablation}) reveals a clear hierarchy of modal dependence. Vision is the most informative single stream (69.97\% accuracy on its own), yet it remains inherently limited by occlusion and semantic aliasing. The integration of time‑series biometrics, treating physiological data as native language tokens rather than disparate sensor readings, provides the critical grounding needed to resolve these ambiguities.

\subsection{Clinical Relevance of Uncertainty}
A critical requirement for deploying AI in veterinary medicine or ethology is the model’s ability to express doubt. Modern large language models are notoriously overconfident, often producing deterministic answers when faced with ambiguous evidence. Our uncertainty quantification experiment demonstrates that Meow‑Omni~1 mitigates this risk.

When presented with congruent data, the model exhibited low predictive entropy ($\mathcal{H} = 1.28$), confidently converging on a single intent. However, when subjected to our adversarial conflict dataset (in which the visual modality suggested a benign state while the biometrics indicated distress), the model’s entropy rose sharply ($\mathcal{H} = 3.15$). This bimodal distribution of probability mass indicates that the model actively weighs the conflicting modalities rather than defaulting to visual heuristics. In a clinical setting, such a high‑entropy state would act as an automated “flag,” allowing the system to defer to a human veterinarian when an animal presents occult pain or concealed distress.

We also include a detailed limitations and future work analysis in Appendix~\ref{app:LF}.

\section{Conclusion} \label{app:conclusion}

Deciphering the intent of non‑verbal species is a deeply complex challenge that has historically been constrained by modality blindness. In this paper, we introduced \textbf{Meow‑Omni~1}, the first Multimodal Large Language Model explicitly engineered for computational ethology to natively co‑embed high‑frequency biological time‑series data with continuous audio‑visual streams.

By performing architectural model surgery and formalising a causal intention‑inference framework, we successfully transitioned from superficial action forecasting to deep intention decoding. Our results on the novel MeowBench suite demonstrate that true semantic understanding of animal behaviour requires grounding observational data in physiological reality. Furthermore, our uncertainty quantification pipeline provides a practical mechanism to detect ambiguous cases, enhancing the model’s safety and interpretability in high‑stakes clinical scenarios.

Ultimately, Meow‑Omni~1 opens a new paradigm for inter‑species communication. By bridging the sensory divide within foundation models, this work lays the technical groundwork for next‑generation veterinary diagnostics, advanced ethological research, and the tech‑enabled conservation of endangered wildlife.

\clearpage
\bibliographystyle{unsrtnat}
\bibliography{mybibliography}

\clearpage
\appendix{Appendix}
\section{Data Processing Pipeline}
\label{appendix:DPP}

\subsection{Bio Dataset Preprocessing}
\label{app:bio_preprocess}
Cat behavioural bio-signal data are generally scarce, particularly datasets that simultaneously provide high temporal resolution and accurate behavioural annotations. To obtain high-quality and consistent signals, we adopt two accelerometer datasets from peer-reviewed studies~\cite{smit2023biodataset1,dunford2024biodataset2}. The original datasets directly annotate each segment with one of the 30 feline intention categories; the full list is available in Appendix~\ref{appedix:label_list}.

In terms of signal processing, the original high-frequency accelerometer data (e.g., 30\,Hz or 60\,Hz) are aggregated into second-level signals by averaging the measurements within each second. This approach reduces high-frequency noise (a common practice in animal behaviour analysis) and mitigates discrepancies in sampling rates across different studies, improving data comparability.

We then construct a task-specific dataset. Invalid or semantically ambiguous labels (e.g., ``other'' categories) are discarded, and samples corresponding to transient intermediate movements that do not clearly represent a stable intention are excluded. The cleaned continuous time series of each individual cat are segmented into fixed-length windows (5, 7, 10, and 15\,seconds). The behavioural label at a future time offset (1, 2, 3, or 5\,seconds ahead) is used as the prediction target, forming a temporal Next‑Behaviour Prediction (NBP) task. We strictly ensure that each time window does not cross individuals or temporal discontinuities, preserving both temporal and individual-level consistency.

Finally, for each sample we generate a user query and a corresponding response via prompt-based generation (see Appendix~\ref{app:prompts}) rather than fixed templates. Three large language models (DeepSeek, GPT, and Gemini) are employed to produce diverse natural-language formulations. The resulting 383,853 labeled TS samples are used for projector alignment in Stage~1 (Section~\ref{sec:training}); a random subset of 2,000 samples is included in Meow‑10K to prevent over‑representation of the TS modality.

\subsection{Video Dataset Preprocessing}
\label{app:video_preprocess}
To extract high-quality, action-centric clips from raw, unannotated cat videos taken from an open-source dataset~\cite{bain2021frozen}, we employ an automated Vision-Language Model (VLM) based preprocessing pipeline. The pipeline uses \textit{Qwen2-VL-72B-Instruct}~\cite{wang2024qwen2} to perform zero-shot frame analysis and consists of four primary stages:

\begin{itemize}
    \item \textbf{Global Action Detection:} The raw video is sampled into coarse frames (up to 16 frames at approximately 1.5\,s intervals). The VLM analyzes these frames to detect significant changes in body posture or position. A gap rescan (approximately 0.5\,s intervals) is applied to temporal gaps exceeding 2.0\,s to avoid missing precursor actions.
    \item \textbf{Precise Temporal Localization:} For each detected anchor, a dense sequence of frames is extracted within a 4.0\,s window at a fine interval of 0.20\,s. The VLM compares consecutive pairs to pinpoint the exact onset frame \(t_{\text{anchor}}\) where the action begins.
    \item \textbf{Verification:} A three-frame verification step compares the subject's state several seconds before the suspected action with frames taken precisely at and shortly after the onset, confirming a definitive state change.
    \item \textbf{Clip Generation and Annotation:} Upon verification, a clip is extracted with a fixed observation window \(T_{\text{obs}} = 6.0\)\,s. Following asymmetric windowing practices in action anticipation~\cite{giancola2021temporally,tai2026action}, the temporal window is heavily biased towards the pre-action phase (spanning from \(t_{\text{anchor}} - 0.85 \times T_{\text{obs}}\) to \(t_{\text{anchor}} + 0.15 \times T_{\text{obs}}\)). Finally, the VLM generates a sequence caption and a specific action label for the cropped segment.
\end{itemize}

Each VLM-generated action label is subsequently mapped to one of the 30 intention classes by a secondary LLM (Qwen3.5-35B). This mapping was reviewed and validated by the same expert groups that curated MeowBench, ensuring consistency with the biological ground truth. All video clips processed in this way are video-only (no accompanying audio track); synchronized audio-video pairs are handled through the separate AudioSet pipeline described below.

\subsection{Audio Dataset Preprocessing}
\label{app:audio_preprocess}
Audio data originate from two independent sources and are processed through distinct pipelines.

\paragraph{AudioSet-Derived Synchronized A/V Clips.}
The first source is AudioSet~\cite{gemmeke2017audio}, a large-scale human-labelled audio event dataset from which we use video clips containing cat vocalisations. Since these clips carry both video and audio streams, excerpt boundaries are inherited directly from the video preprocessing stage, ensuring temporal alignment by construction.

Each extracted audio track is first verified for cat sound presence using an Audio Spectrogram Transformer (AST) classifier fine-tuned on AudioSet~\cite{gong2021ast}. The classifier scores each clip against cat-related AudioSet labels (\textit{cat}, \textit{meow}, \textit{purr}, \textit{caterwaul}). Clips scoring above 0.10 are retained; clips below 0.03 are discarded. Marginal clips (0.03--0.10) undergo stationary noise reduction with a suppression factor of 0.85, after which the classifier is re-applied; a clip is retained only if the post-denoising score meets the 0.03 threshold.

Verified clips are then passed to Qwen2.5-VL-7B~\cite{bai2025qwen25vltechnicalreport}, which jointly analyses the video and audio to produce an audio-focused caption and an action label. The model is explicitly instructed to use visual context only as interpretive reference, grounding the final description and classification in the acoustic content. Simultaneously, the model assigns a score from 1 to 10 to each of the 30 intention classes; scores are normalized to form a probability distribution used as the training target for these samples.

\paragraph{Standalone Audio Clips.}
The second source consists of standalone audio recordings from Freesound~\cite{fonseca2017freesound} and the huggingface/taozi555/cat\_class dataset~\cite{taojiang_2026_cat_class}. These carry short human-written captions but no associated video. No sound verification step is required; the original captions serve as the starting point.

Processing is carried out via a text-only pipeline using Qwen3.5-35B-A3B-GPTQ-Int4~\cite{qwen3.5}, served locally or through an OpenAI-compatible API. The original caption is expanded into a behaviour-relevant semantic description, grounded in the source text, with uncertain inferences phrased tentatively. Importantly, no intention label is assigned; the expanded caption itself serves as the training target. This approach means the model is trained \emph{not} to directly output an intention class from standalone audio, but to produce a faithful description of the acoustic behaviour, from which latent intent can later be inferred when combined with other modalities.

\subsection{Meow‑10K Dataset Assembly}
\label{app:meow10k_assembly}
All samples produced by the preceding pipelines are consolidated into the Meow‑10K training set. To prevent the abundant TS‑only data from overwhelming the other modalities, we randomly subsample 2,000 sequences from the full 383,853 TS pool. The remaining components comprise:
\begin{itemize}
    \item Video‑only clips from the Bain dataset (Section~\ref{app:video_preprocess});
    \item Naturally synchronized audio‑video pairs from AudioSet (Section~\ref{app:audio_preprocess});
    \item Standalone audio captions from Freesound and cat\_class (Section~\ref{app:audio_preprocess});
    \item A small set of expert‑verified synthetic quad‑modal samples, constructed following the MeowBench protocol (Section~\ref{sec:meowbench}), where a synchronized A/V pair is matched with a TS sample sharing the same intention label and the combination is verified by one of the expert groups.
\end{itemize}
The final Meow‑10K dataset contains 10,831 samples, each consisting of a natural-language query‑response pair and, where applicable, an intention label drawn from the unified 30‑class taxonomy.

\section{Baselines Detailed Discussion}
\label{appendix:baselines}

\subsection{Video Baseline}
We implemented a SOTA video baseline using the Qwen3.5‑122B‑A10B large vision‑language model. The evaluation was conducted in a zero‑shot, audio‑blind setting with the model’s internal “thinking” module disabled to strictly isolate and assess native visual‑temporal reasoning capabilities. The visual preprocessing pipeline included uniform temporal sampling of 8 frames per input clip, which were provided as sequential inputs to the model. The baseline was evaluated on MeowBench to benchmark the model’s visual understanding against ground‑truth annotations. During evaluation, each model response was processed through a regular‑expression‑based parsing logic to extract the predicted multiple‑choice letter (A, B, C, or D), which was then mapped to the corresponding benchmark answer. Accuracy was employed as the primary evaluation metric to ensure parity with the audio baseline.

\subsection{Biosignal Baseline}
The biosignal baseline was implemented to evaluate how much intention‑related information can be extracted from the TS modality alone. We based this baseline on the cat and dog IMU behaviour recognition model proposed in \cite{Chen_2025}, and adapted it to our benchmark format. Since the original work was designed for a different data setting, our implementation should be regarded as a paper‑style reproduction rather than a full reproduction of the original experiment. In particular, we retained the core temporal modelling idea of combining convolutional feature extraction with recurrent sequence modelling, while simplifying the architecture to fit the structure of our benchmark data.

\begin{figure}[htbp]
    \centering
    \includegraphics[width=0.3\textwidth]{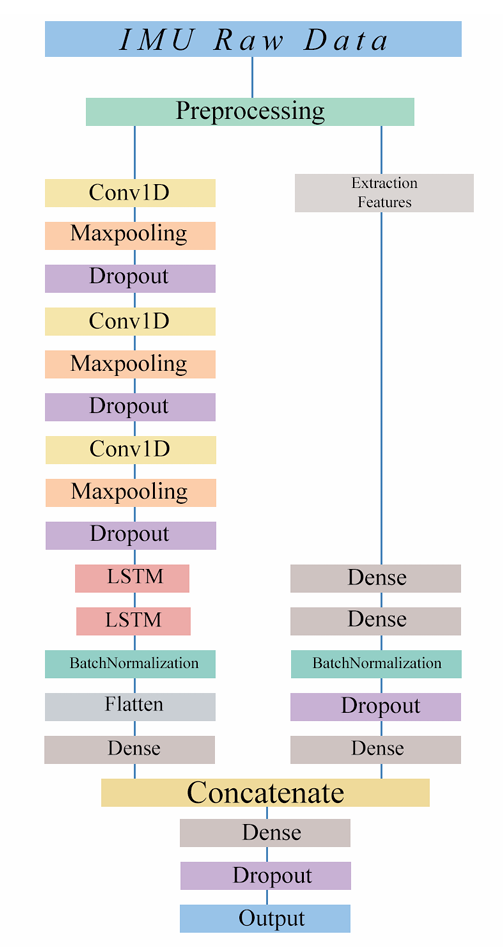}
    \caption{Baseline for biosignal modality, adapted from \cite{Chen_2025}.}
    \label{fig:bioBaseline}
\end{figure}

The final baseline architecture follows the main design philosophy of \cite{Chen_2025}. First, the $(5,3)$ input tensor is transposed to channel‑first format and passed through two 1D convolution layers. The first convolution layer uses 32 channels, followed by ReLU activation, max pooling, and dropout. The second convolution layer increases the feature dimension to 64 channels, followed again by ReLU and dropout. The output sequence is then transposed back to time‑major format and fed into a single‑layer LSTM with hidden size 64. The hidden state of the last time step is batch‑normalized and passed to a two‑layer fully connected classifier with dropout, producing the final logits over the 10 intention classes. This design preserves the paper’s idea of combining local temporal pattern extraction and sequential modelling, while remaining lightweight enough for our short benchmark sequences. Figure~\ref{fig:bioBaseline} shows the architecture of the baseline; the main difference from the original is that we omit the manually extracted feature branch (right‑hand side of the original figure).

For training, the benchmark data were split into train, validation, and test sets with a ratio of approximately 70:10:20 using stratified sampling over the intention labels. The model was optimized with Adam using a learning rate of $10^{-4}$ and a batch size of 64. To avoid selecting the final model based on the test set, we adopted early stopping using validation accuracy only. The checkpoint with the best validation accuracy was saved, and the final test accuracy was computed once using this checkpoint after training stopped. In addition to the classification accuracy on the intention labels, we also evaluated the model in the original multiple‑choice benchmark format by mapping the predicted intention class back to the corresponding answer option (A/B/C/D) for each sample. Raw prediction outputs, including predicted labels and logits, were stored for later benchmark analysis and potential recalculation if problematic items were removed.

This baseline serves two purposes in our study. First, it provides a uni‑modal reference point for the biosignal modality, allowing us to estimate how much behavioural intention can be inferred from short acceleration signals alone. Second, it establishes a practical bridge between prior wearable‑sensor behaviour recognition work and our multimodal MLLM benchmark, showing how an existing IMU‑based architecture can be adapted to a new intention classification setting.

\subsection{Audio Baseline}
We implemented three top‑performance audio baselines from Ntalampiras et al.~\cite{ntalampiras2019automatic}: a support vector machine (SVM), a directed acyclic graph hidden Markov model (DAG‑HMM), and a class‑specific hidden Markov model. All methods used the same acoustic preprocessing pipeline, including silence removal, mel‑frequency cepstral coefficient (MFCC)‑based features, and features describing temporal variation in the signal. The baselines were trained on our 22‑class audio training set and evaluated on MeowBench in the audio‑only setting.

MeowBench contains 30 answer‑option labels, whereas our audio‑only training set covers 22 classes that correspond to audible behaviours. The remaining eight labels are primarily pose‑, motion‑, scene‑, or sensor‑oriented categories, which fall outside the audio‑only supervision space; for those eight classes, audio‑based prediction was not attempted. During evaluation, each model predicted one class label from the 22‑class set for an input clip; when the gold label belonged to the 22 audible classes, the predicted label was mapped to the corresponding answer option in the multiple‑choice benchmark. Accuracy over the class‑matched subset was used as the main evaluation metric.

\subsection{Omni Baseline}
We evaluate \textbf{Qwen3.5‑Omni‑Plus}~\cite{qwen3omni2025}, accessed via the DashScope API, as our omni‑modal baseline. Qwen3.5‑Omni‑Plus is a proprietary instruction‑following variant built upon the Qwen3‑Omni architecture~\cite{qwen3omni2025}, a Thinker–Talker Mixture‑of‑Experts end‑to‑end multimodal model that natively processes text, images, audio, and video within a single unified model. We use Qwen3.5‑Omni‑Plus rather than the open‑weight Qwen3‑Omni directly for two practical reasons: the open‑weight Qwen3‑Omni release does not provide a publicly accessible instruction‑following API, and its Captioner variant does not support video input, making both unsuitable for our quad‑modal evaluation format. Qwen3‑Omni achieves open‑source SOTA performance on 32 of 36 audio and audio‑visual benchmarks, surpassing closed‑source systems including Gemini 2.5 Pro and GPT‑4o‑Transcribe~\cite{qwen3omni2025}, establishing it as a strong zero‑shot upper‑bound comparator.

\paragraph{Prompting strategy}
The model receives video and audio as native modality inputs. As Qwen3.5‑Omni‑Plus does not support raw TS as a modality, bio‑signal data is injected as a structured textual summary computed from the accelerometer \texttt{.npy} array, reporting array shape, mean absolute magnitude, average temporal change, and per‑channel mean, standard deviation, minimum, and maximum across the first eight channels. This summary is prepended to the MeowBench MCQ question. The system prompt instructs the model to treat all three inputs as co‑equal evidence and respond with a single letter on the first line followed by exactly two explanatory sentences. Each MCQ is formulated as a \emph{structured decision problem}, where the model must select one option from four candidates (A/B/C/D), each corresponding to a semantically distinct hypothesis about the underlying behaviour or cross‑modal relationship.

\paragraph{Ablation conditions}
To isolate each modality’s contribution, we evaluate the same backbone under four input configurations: \textit{Video + Audio}, \textit{Video + Bio}, \textit{Audio + Bio}, and \textit{Video + Audio + Bio}. All conditions use greedy decoding (temperature $= 0$), ensuring that performance differences are attributable solely to modality availability rather than model or sampling variation.

\section{Limitations and Future Work} \label{app:LF}
Despite its strong performance, this study has several limitations that present opportunities for future research. 

First, our NBP strategy relies on the assumption that an animal's internal intent will manifest as a physical action in the immediate temporal horizon. This proxy may fail to capture protracted intentions, such as a predatory stalking state that persists for extended periods without a behavioural transition. Future work will explore expanding the temporal receptive field and integrating longer contextual windows to capture these multi-stage intent states.

Second, the MeowBench conflict dataset relies on intent-matched synthesis. While the physiological plausibility of these synthesized multi-modal triplets was verified by expert ethologists, the dataset remains inherently synthetic. Future verification must be conducted on entirely native, independently collected, and isolated data from real-world environments to ensure robustness against environmental noise and true out-of-distribution scenarios. 

Third, Meow-Omni 1 currently operates as a passive reasoning engine. To enhance its utility as a real-time auditor in wildlife protection, future iterations will explore a \textbf{Full-Duplex architecture} with integrated \textbf{instant Text-to-Speech (TTS)} capabilities. This would enable the model to not only observe but actively interface with its environment, providing immediate auditory alerts or feedback based on inferred states.

Finally, while our framework was explicitly designed using \textit{Felis catus} as a scalable template, cross-species generalisation remains an open question. Future work will focus on zero-shot and few-shot transfer learning, investigating whether the physiological-behavioural alignments learned in domestic felines can map to endangered feline taxa, such as the Amur Leopard (\textit{Panthera pardus orientalis}).

\section{Prompt Design for Query Generation}
\label{app:prompts}

We design four types of prompts to generate natural language queries for the TS prediction task. 
Here, $A$ denotes the length of the input time window, and $B$ denotes the prediction horizon (i.e., $B$ seconds after the window).

\textbf{(1) Prediction with window length $A$}
\begin{tcolorbox}
Generate 300 different natural language queries asking to predict the cat’s behavior after a time window of \{A\} seconds.\\

Each query must include "\texttt{<|ts\_start|><|ts\_unit|><|ts\_end|>}".\\

Example (keep "\{A\}" unchanged):\\
"Given a \{A\}-second window \texttt{<|ts\_start|><|ts\_unit|><|ts\_end|>}, predict the behavior."\\

Encourage diverse wording; output a Python list of strings.
\end{tcolorbox}

\textbf{(2) Prediction with delay $B$}
\begin{tcolorbox}
Generate 300 different natural language queries asking to predict the cat’s behavior \{B\} seconds after a given TS segment.\\

Each query must include "\texttt{<|ts\_start|><|ts\_unit|><|ts\_end|>}".\\

Example (keep "\{B\}" unchanged):\\
"Given \texttt{<|ts\_start|><|ts\_unit|><|ts\_end|>}, predict the behavior after \{B\} seconds."\\

Encourage diverse wording; output a Python list of strings.
\end{tcolorbox}

\textbf{(3) Prediction with window $A$ and delay $B$}
\begin{tcolorbox}
Generate 300 different natural language queries asking to predict the cat’s behavior \{B\} seconds after a time window of \{A\} seconds.\\

Each query must include "\texttt{<|ts\_start|><|ts\_unit|><|ts\_end|>}".\\

Example (keep "\{A\}" and "\{B\}" unchanged):\\
"Given a \{A\}-second window \texttt{<|ts\_start|><|ts\_unit|><|ts\_end|>}, predict the behavior after \{B\} seconds."\\

Encourage diverse wording; output a Python list of strings.
\end{tcolorbox}

\textbf{(4) Basic prediction without temporal parameters}
\begin{tcolorbox}
Generate 300 different natural language queries asking to predict the cat’s behavior from a given TS segment.\\

Each query must include "\texttt{<|ts\_start|><|ts\_unit|><|ts\_end|>}".\\
Example:\\
"Given \texttt{<|ts\_start|><|ts\_unit|><|ts\_end|>}, predict the cat’s behavior."\\

Encourage diverse wording; output a Python list of strings.
\end{tcolorbox}

\textbf{(5) Response generation from feature summaries}

\begin{tcolorbox}
Generate 300 different natural language assistant responses for a cat behavior prediction task.\\

You are given:\\
- a feature summary describing the acceleration pattern of one behavior class: \{features\}\\
- the target behavior label: \{label\}\\

Write one-sentence responses that infer the behavior from the described signal pattern. The response should explain the prediction briefly rather than only repeating the label.\\

Example:\\
"The recurring peaks in the acceleration signal confirm the cat is likely walking forward."\\

Encourage diverse wording; output a Python list of strings.
\end{tcolorbox}

\section{Feature Construction} \label{appedix:label_list}
The feature summaries provided in the prompt are constructed based on the statistical characteristics of the acceleration data for each behaviour class. 
For each class, we analyse aggregated signals and derive representative patterns in terms of variance, temporal dynamics, and overall motion structure. 
These patterns are then expressed in natural language and used as the \{features\} input to guide response generation.

The behaviour-specific feature descriptions used in this work are as follows:

\begin{itemize}
\item Feed: shows low variance and small changes over time, with acceleration concentrated around a steady downward Z component.
\item Groom: exhibits moderate variance and noticeable fluctuations, especially along the Y axis, indicating active but controlled motion.
\item Rest: has a wide range with occasional spikes but generally low temporal variation, reflecting mostly static behavior with intermittent disturbances.
\item Run: characterized by high variance and large step-to-step changes, indicating fast and continuously changing acceleration patterns.
\item Shake: displays moderate variance with consistently negative offsets and rapid short-term changes, suggesting oscillatory motion.
\item Trot: shows pronounced variance and rhythmic changes over time, indicating structured and repeating acceleration patterns.
\item Walk: features moderate variance with smooth and regular temporal changes, reflecting steady and periodic motion.
\item active\_climbing: shows sustained multidirectional movement, with noticeable variation on all three axes and relatively frequent changes over time.
\item active\_jumping.horizontal: exhibits strong short-term fluctuations across all axes, with acceleration centered near zero and clear burst-like motion.
\item active\_jumping.vertical: is characterized by large vertical excursions and rapid second-to-second changes, indicating repeated impulsive movement.
\item active\_playfight.fighting: displays irregular and active motion with substantial frame-to-frame changes, especially along the X and Z axes.
\item active\_playfight.playing: contains only a single aggregated sample, so it appears as a fixed acceleration state with no observable temporal variation.
\item active\_rubbing: shows moderate multidirectional variation with a wider spread on Z and relatively smooth second-to-second change.
\item active\_trotting: presents rhythmic movement with moderate spread on all axes and a stable pattern of repeated variation over time.
\item active\_walking: shows regular periodic motion with moderate dispersion and smooth temporal transitions in acceleration.
\item inactive\_lying.crouch: has a stable posture-like pattern with very small second-to-second changes and relatively fixed acceleration orientation.
\item inactive\_lying.down: shows low temporal variation with a consistent offset in acceleration, reflecting a mostly still posture.
\item inactive\_lying.resting: is highly stable, with tightly clustered values on all axes and minimal change over time.
\item inactive\_sitting.down: displays moderate spread in acceleration with noticeable but not abrupt temporal changes.
\item inactive\_sitting.stationary: maintains a very stable pattern over time, with clear orientation offsets but very small second-to-second variation.
\item inactive\_sitting.up: shows a relatively steady posture with moderate spread and gentle temporal fluctuations.
\item inactive\_standing.stationary: has a largely stable acceleration pattern with posture-dependent offsets and limited temporal movement.
\item inactive\_standing.up: shows marked short-term movement with substantial second-to-second change despite being labeled as a standing transition state.
\item maintenance\_grooming: exhibits moderate multidirectional spread with smooth temporal evolution and no strong single-axis dominance.
\item maintenance\_littering.defecating: shows a fairly steady pattern with limited temporal change and a consistent downward Z tendency.
\item maintenance\_littering.digging: displays repeated controlled movement with moderate spread and steady changes across all three axes.
\item maintenance\_littering.none: shows a compact acceleration pattern with mild fluctuations and a stable overall orientation.
\item maintenance\_littering.urinating: has a relatively stable pattern with moderate spread and modest temporal change, especially on Z.
\item maintenance\_nutrition.eating: shows a steady acceleration orientation with limited second-to-second variation and moderate axis spread.
\item maintenance\_scratching: exhibits active multidirectional movement with noticeable spread and frequent short-term changes.
\item maintenance\_shake.body: shows strong whole-body oscillatory motion with large spread and rapid temporal variation, especially on Z.
\item maintenance\_shake.head: displays sharp and frequent short-term changes, with pronounced oscillation and strong variability across axes.
\item other\_social.allogrooming: exhibits moderate controlled movement with a stable Y component and gentle temporal variation.
\end{itemize}

\noindent For samples that do not conform well to the characteristic patterns of their assigned class, or that correspond to transitional states where the behaviour is about to change, we apply separate matching strategies to generate appropriate responses.

\section{Reproducibility Details}
\label{app:repro}

We provide all information necessary to reproduce the main experimental results of Meow‑Omni~1.
The model architecture, training pipeline, and evaluation protocol are fully described in the main paper;
here we consolidate the exact implementation details.

\subsection*{Hardware and Software Environment}
All experiments were conducted on a single computing node equipped with
\textbf{8$\times$ NVIDIA H200 80\,GB GPUs} (NVLink interconnected).

\subsection*{Model Architecture}
Meow‑Omni~1 is based on the MiniCPM‑V backbone.
We expand the tokenizer with three special tokens
\texttt{<|ts\_start|>}, \texttt{<|ts\_unit|>}, and \texttt{<|ts\_end|>}
and integrate a time‑series encoder adapted from Intern‑S1 Pro.
A linear projector maps the TS encoder outputs to the LLM’s hidden dimension.
During the forward pass, TS embeddings are interleaved with visual tokens and text tokens
in a unified multimodal sequence.
The TS encoder itself is frozen during all training stages.

\subsection*{Training Pipeline}
Training proceeds in two stages.

\subsubsection*{Stage~1: Projector Alignment}
\begin{itemize}
    \item \textbf{Data:} 383,853 time‑series samples (window lengths 5, 7, 10, 15\,s)
          paired with behaviour labels.
    \item \textbf{Trainable parameters:} Only the TS projector; the LLM backbone and TS encoder are frozen.
    \item \textbf{Optimization:} AdamW,
          learning rate $10^{-4}$,
          weight decay $0.0$,
          batch size $1$ per device,
          gradient accumulation steps $2$.
    \item \textbf{Learning rate schedule:} Cosine annealing with linear warmup for 3\% of total steps;
          maximum number of epochs $10$.
    \item \textbf{Early stopping:} Validation accuracy monitored; patience $1$ epoch.
          The best checkpoint (epoch~2) is saved as \textit{Meow‑Omni~1 Aligned}.
    \item \textbf{Loss:} Cross‑entropy on the predicted next behaviour class.
\end{itemize}

\subsubsection*{Stage~2: Multimodal Specialization}
\begin{itemize}
    \item \textbf{Data:} 10,831 high‑quality quad‑modal samples from the Meow‑10K dataset,
          including partial‑modality combinations (V+A, V+TS, TS+A, V+A+TS).
    \item \textbf{Trainable parameters:} LLM backbone only; all encoders and projectors frozen.
    \item \textbf{Optimization:} AdamW as above,
          learning rate $2\times10^{-5}$,
          weight decay $0.0$,
          batch size $1$ per device,
          gradient accumulation steps $2$.
    \item \textbf{Learning rate schedule:} Cosine decay with 3\% warmup;
          maximum epochs $5$.
    \item \textbf{Early stopping:} Validation accuracy used; patience $1$ epoch.
          The final \textit{Meow‑Omni~1} checkpoint was taken at epoch~1.
    \item \textbf{Loss:} Next‑token prediction loss over the full multimodal sequence,
          including special tokens.
\end{itemize}

\subsection*{Evaluation Procedure}
All MeowBench results are obtained with a single deterministic greedy decoding pass
(temperature $T=0$, top‑p disabled).
The predicted intent is mapped back to the multi‑choice option (A/B/C/D) via a regex that extracts
the first capital letter appearing after the model’s answer prefix.

\section{Potential Negative Impacts}
Meow‑Omni 1 is designed to decode feline behavioural intent from multimodal sensor data. The technology could be misused for continuous behavioural surveillance, e.g., by employers or insurance companies to profile pet health or owner behaviour. Misinterpretation of ambiguous states could lead to incorrect veterinary decisions if used without human oversight. To mitigate these risks, we advocate for (1) keeping the human in the loop, (2) deploying deterministic uncertainty flags (as demonstrated in our uncertainty quantification experiments) to defer ambiguous cases, and (3) restricting high‑stakes decision‑making to certified veterinary professionals only. We will also release the model under a license that requires attribution and prohibits unlicensed commercial use.

\section{Licenses of Existing Assets}
All third‑party datasets, pre‑trained models, and code packages used in this work are publicly released under permissive licenses (CC‑BY 4.0 or equivalent). The specific license for each asset is noted in its original publication and, where applicable, in the asset metadata. No copyrighted material has been used without permission.

% \newpage
% \input{checklist.tex}

\end{document}